\documentclass{article}

\usepackage[preprint]{neurips_2024}

\usepackage[utf8]{inputenc} % allow utf-8 input
\usepackage[T1]{fontenc}    % use 8-bit T1 fonts
\usepackage{hyperref}       % hyperlinks
\usepackage{url}            % simple URL typesetting
\usepackage{booktabs}       % professional-quality tables
\usepackage{amsfonts}       % blackboard math symbols
\usepackage{float}          % for [H] float placement
\usepackage{nicefrac}       % compact symbols for 1/2, etc.
\usepackage{microtype}      % microtypography
\usepackage{xcolor}         % colors
\usepackage{comment}
\usepackage{natbib}
\usepackage{multirow}

\title{MERaLiON-GR: Speech Gender Recognition Model for English and SEA Languages}

\author{
  MERaLiON Team \\
  Institute of Advanced Intelligence and Computing (IAIC), A*STAR, Singapore \\
  Corresponding Authors:%\\
  Qiongqiong Wang (\texttt{wang\_qiongqiong@a-star.edu.sg})%\\
}

\usepackage{booktabs,siunitx,adjustbox}
\usepackage{makecell} % only for multi-line model names if needed
\begin{document}

\maketitle

\begin{abstract}
We present \textbf{MERaLiON-GR}\footnote{The MERaLiON-GR-v1 model: \url{https://huggingface.co/MERaLiON/MERaLiON-GR-v1}. An online demo: \url{https://meralion.org/analysis}.}
, a speech gender recognition system that performs binary classification (female / male) on English and Southeast Asian (SEA) languages. The model finetunes MERaLiON-SpeechEncoder-2, a large conformer-based transformer pre-trained on a broad speech corpus, and applies parameter-efficient fine-tuning via Low-Rank Adaptation (LoRA) to adapt the encoder to the gender recognition task, and appends a multi-scale ECAPA-TDNN downstream network with attention pooling and a lightweight linear classifier. Extensive evaluations across multilingual Singaporean and Southeast Asian languages (English, Chinese, Malay, Tamil, Thai, Vietnamese, Indonesian, and Khmer) show that MERaLiON-GR consistently surpasses the state-of-the-art gender recognition model Vox-Profile and a large Audio-LLM, in both full-utterance and segment-level evaluation modes. The results underscore the value of dedicated speech models in achieving accurate paralinguistic understanding and strong cross-lingual generalization.
\end{abstract}

\section{Introduction}

Gender recognition from speech is a fundamental paralinguistic task with applications ranging from speaker diarization and speaker profiling to personalised speech interfaces and audio forensics. While human listeners can reliably infer speaker gender from brief utterances, automatic systems must generalize across diverse recording conditions, languages, accents, age groups, and channel characteristics.

In contrast to earlier approaches based on handcrafted acoustic features, such as fundamental frequency (F0), formants, and spectral descriptors, modern deep learning methods learn discriminative representations directly from speech signals. These methods typically employ convolutional, recurrent, ECAPA-TDNN~\citep{ECAPA}, or Transformer-based architectures operating on raw waveforms or log-Mel filterbank features. More recently, self-supervised speech models, including HuBERT~\citep{hsu2021hubert},  wav2vec 2.0~\citep{baevski2020wav2vec},  and WavLM~\citep{chen2022wavlm}, have significantly advanced speech representation learning. Their pretrained representations encode rich acoustic, speaker, and linguistic information, enabling substantial improvements across downstream paralinguistic tasks, including gender recognition.

In this work, we built a complete pipeline for multilingual speech gender recognition based on a large pre-trained conformer encoder, MERaLiON-SpeechEncoder-2\footnote{The MERaLiON-SpeechEncoder-2 model: \url{https://huggingface.co/MERaLiON/MERaLiON-SpeechEncoder-2}}~\citep{huzaifah2024speechfoundationmodelsingapore}. MERaLiON-SpeechEncoder-2 is a large conformer-based transformer trained with a Best-RQ objective~\citep{chiu2022selfsupervised} on a multilingual corpus that includes a significant proportion of Singapore English and other Southeast Asian languages, making it particularly well suited for gender recognition benchmarks centred on Southeast Asia. However, fully fine-tuning all 24 layers is computationally expensive and may increase the risk of catastrophic forgetting. To address this, we adopt Low-Rank Adaptation (LoRA)~\citep{hu2022lora}, which parameterizes weight updates in the attention projection layers using low-rank factors rather than updating the full weight matrices. This yields a parameter-efficient adaptation strategy for the downstream task.

The main contributions of this work are: (1) a parameter-efficient multilingual GR model built on MERaLiON-SpeechEncoder-2 with LoRA and ECAPA-TDNN, outperforming Vox-Profile~\citep{feng2025vox} on 12 of 15 public benchmarks spanning eight languages across Southeast Asia and English; (2) robust segment-level recognition on short in-the-wild speech, with gains of up to 4.32\,pp over Vox-Profile on Singapore language sets; and (3) a demonstration that injecting GR estimates as paralinguistic metadata substantially improves Audio-LLM performance on gender-related question answering, consistent with our prior findings on contextual paralinguistic understanding in Audio-LLMs~\citep{cpqa_asru}.

% ─── 4. Model Architecture ──────────────────────────────────────────────────
\section{Model description}
\label{model_description}

The overall architecture consists of three components: (1) the pre-trained MERaLiON-SpeechEncoder-2 backbone~\citep{huzaifah2024speechfoundationmodelsingapore}, (2) a layer-attentive aggregation module, and (3) an ECAPA-TDNN downstream network followed by a linear classification head.

\subsection{Backbone: MERaLiON-SpeechEncoder-2 with LoRA}
MERaLiON-SpeechEncoder-2 is a Conformer-based transformer~\citep{huzaifah2024speechfoundationmodelsingapore}. To enable parameter-efficient adaptation while reducing computational cost and the risk of overfitting, we insert Low-Rank Adaptation (LoRA)~\citep{hu2022lora} adapters into the attention projection layers. The LoRA rank and the scaling factor are both set to 16. We further adopt rsLoRA normalization~\citep{kalajdzievski2023rslora} and apply a LoRA dropout rate of 0.05.

All 25 hidden-state outputs (the input embeddings plus all 24 transformer layers) are retained. A learned soft-attention mechanism over per-layer mean representations produces a weighted combination $H = \sum_{l} a_l h_l$, capturing complementary phonetic and speaker-level cues from different layers.

\subsection{Downstream Model: ECAPA-TDNN}

The aggregated representation is processed by an ECAPA-TDNN network adapted for gender recognition~\citep{ECAPA}. It is projected to 512 dimensions and passed through three SE-Res2Net blocks (dilations $\{1,2,3\}$, SE reduction factor 8); All BatchNorm layers are replaced with GroupNorm layers that are in the original ECAPA-TDNN structure. Attention pooling aggregates temporal frames into a fixed-length embedding $\mathbf{e}$, which is projected to 256 dimensions and classified by a linear head with RMSNorm and GELU activation. The classification head maps the 256-dimensional embedding to two class logits:
\begin{equation}
    \hat{y} = W_2 \, g\!\left(\mathrm{RMSNorm}(W_1 \mathbf{e})\right)
\end{equation}
where $g(\cdot)$ denotes the GELU activation.

%\footnote{As in this example.}

\section{Training Details}

\subsection{Training datasets}
\label{sec:train_data}
The training data consists of six dataset partitions: VoxCeleb1~\citep{nagrani17_interspeech} (gender-labelled) and IMDA PART1–5 (Singaporean English).

\begin{itemize}
    \item \textbf{VoxCeleb1:} Segments drawn from the VoxCeleb1 corpus. 
    \item \textbf{IMDA PART1--5:}   
    We use five partitions of a large-scale Singapore speech corpus developed by the Info-communications Media Development Authority (IMDA) of Singapore and derived from the National Speech Corpus (NSC)~\citep{koh19_interspeech,wang2025advancing}. They cover a broad range of speaking conditions: prompted phonetically-balanced scripts with local accents (PART1); prompted sentences drawn from everyday topics such as food, people, and locations, rich in local terms (PART2); spontaneous conversational speech on daily-life topics (PART3); code-switching between Singapore English and Mandarin, Malay, or Tamil (PART4); and stylised speech including debates, finance discussions, and emotional expressions (PART5). This diversity of speaking styles, accents, and code-switching patterns makes the NSC particularly valuable for training a robust multilingual Gender recognition model targeting Southeast Asian speech.
\end{itemize}

\subsection{Training Configuration}
\label{sec:wce_loss}
A multi-dataset training strategy was adopted across the six training partitions described in Section~\ref{sec:train_data}. To ensure balanced optimization across datasets of different sizes, each training batch consisted of 32 samples drawn from a single dataset, where the dataset was sampled with probability proportional to its size. Class imbalance was addressed using the class-balanced weighted cross-entropy loss described in Section~\ref{sec:wce_loss}. To improve model robustness without incurring the inference overhead of model ensembling, the final model parameters were obtained by averaging the weights of the four checkpoints with the lowest categorical loss on the development set.

Optimization employed two parameter groups: the LoRA adapters inserted into the frozen backbone were trained with a learning rate of ($5\times10^{-5}$) and a weight decay of ($5\times10^{-4}$), while the downstream classification network used a learning rate of ($6\times10^{-4}$) and a weight decay of ($8\times10^{-5}$). A cosine learning-rate scheduler with a linear warm-up ratio of 0.08 was applied throughout training. Label smoothing with ($\epsilon = 0.1)$ was used to improve generalization. The model was trained for 15 epochs on a single NVIDIA H100 GPU, using the development-set categorical loss as the early stopping criterion.

To address class imbalance, we optimize the model using a weighted cross-entropy loss for gender classification:
\begin{equation}
\mathcal{L}_{\text{CE}} = - \sum_{i=1}^{C} w_i \cdot y_i \log(\hat{y_i}),
\end{equation}
where \(C\) denotes the number of gender classes, \(y_i\) and \(\hat{y_i}\)  are the ground truth and the predicted probability for the \(i\)-th class, respectively. \(w_i\) is the weight assigned to the \(i\)-th class to compensate for class imbalance. The per-class weights $w_i$ are computed from the effective number of training samples per class~\citep{cui2019classbalanced}.

%Early stopping was triggered if validation performance did not improve for three consecutive e

\section{Evaluation Setup}
\label{sec:datasets}

\subsection{Evaluation Datasets}
Model performance was assessed using both manually curated Singapore speech datasets and publicly available multilingual benchmarks, providing evaluation across a wide range of languages and recording conditions.

\subsubsection{Manually Curated Singapore Evaluation Set}

The Singapore evaluation set (SG-ECMT) comprises speech samples in the country's four major languages: Singlish, Mandarin Chinese, Malay, and Tamil. The data was derived from our proprietary unlabeled speech corpus and consists of 10--30 second in-the-wild speech segments. Following the data processing pipeline proposed in \citep{cpqa_interspeech}, speech segments were selected from a large collection of in-the-wild recordings based on their paralinguistic metadata~\citep{wang2025benchmarking}. Gender labels were automatically estimated using a WavLM-ECAPA\footnote{\url{https://github.com/wenet-e2e/wespeaker}} model~\citep{chen2022wavlm,ECAPA} fine-tuned on the VoxCeleb2 dataset~\citep{chung2018voxceleb2}. Inference was performed using a 3-second sliding window with a 1-second overlap (i.e., a 2-second hop), producing one gender prediction every two seconds. The predicted gender labels were subsequently reviewed and corrected by native-speaking annotators for each language to ensure annotation quality. The final evaluation set comprises 466 Singlish, 466 Chinese, 479 Malay, and 469 Tamil speech samples.

\subsubsection{Public evaluation set}

To evaluate generalization, we leverage a diverse collection of public benchmarks spanning eight languages: English (FLEURS~\citep{conneau2023fleurs}, IEMOCAP~\citep{busso2008iemocap}, and Common Voice~\citep{ardila2020common}), Chinese (Common Voice~\citep{ardila2020common}), Malay (SMALDUSC~\citep{magichub2023smaldusc}), Tamil (OpenSLR SLR65~\citep{he2020open}, EmoTa~\citep{thevakumar2025emota}, and Common Voice~\citep{ardila2020common}), Thai (THAI-SER~\citep{wongpithayadisai2025thaiser}, Thai Elderly Speech~\citep{datawow2023thaielderly}, and Common Voice~\citep{ardila2020common}), Vietnamese (Common Voice~\citep{ardila2020common}), Indonesian (IndoWaveSentiment~\citep{bustamin2024indowave} and Common Voice~\citep{ardila2020common}), and Khmer (FLEURS~\citep{conneau2023fleurs}).

\section{Baselines and Metrics}

We benchmark MERaLiON-GR against two representative baselines. Vox-Profile~\citep{feng2025vox} is a speech foundation model designed to characterize diverse speaker and speech attributes and represents the current state of the art in standalone gender recognition. MERaLiON-v2\footnote{The MERaLiON-2-10B model: \url{https://huggingface.co/MERaLiON/MERaLiON-2-10B}} is a general-purpose Audio-LLM built on the MERaLiON-AudioLLM framework~\citep{meralion}, capable of performing natural-language instruction following over speech input. We include it as a representative Audio-LLM baseline to evaluate whether a general-purpose multimodal model can perform fine-grained paralinguistic recognition without task-specific training.

For the standalone gender recognition models (Vox-Profile and the proposed MERaLiON-GR), performance is evaluated using classification accuracy. To ensure a fair comparison, MERaLiON-v2 is prompted with a fixed instruction:

\begin{quote}
\texttt{PROMPT = "Determine the speaker's gender in the given audio. Reply with a single label from: Female, Male"}
\end{quote}

The generated response is mapped to one of the two gender labels, and classification accuracy is computed using the same evaluation protocol as the standalone gender recognition models.

\section{Results and Discussion}
\label{sec:results}

\subsection{Public Benchmark Results}
Table~\ref{tab:public} reports gender recognition (GR) accuracy on the public multilingual benchmarks. We compare the proposed MERaLiON-GR model with the state-of-the-art standalone GR model, Vox-Profile, and the general-purpose Audio-LLM, MERaLiON-v2. Overall, MERaLiON-v2 consistently underperforms the dedicated GR models, indicating that a general-purpose Audio-LLM is less effective for fine-grained gender recognition without task-specific optimization.

Compared with Vox-Profile, MERaLiON-GR achieves higher accuracy on 12 of the 15 evaluated test sets and matches its performance on the Khmer FLEURS benchmark. The proposed model attains perfect accuracy (100.00\%) on the English FLEURS, Tamil OpenSLR, and Thai Elderly benchmarks. The largest improvements are observed on Tamil EmoTa (+4.71,pp), Vietnamese Common Voice (+3.14,pp), and Indonesian IndoWave (+3.00,pp), demonstrating strong generalization across diverse languages and recording conditions. Performance decreases are observed only on the Malay SMALDUSC dataset ( $-4.40$,pp) and the Thai SER dataset ($-2.09$,pp). These degradations are likely due to domain mismatch, as both datasets differ substantially from the training data in terms of recording conditions and speaking style.

Table~\ref{tab:meta_audiollm} investigates whether incorporating the predicted gender from MERaLiON-GR can improve Audio-LLM performance. Specifically, the predicted gender is injected as paralinguistic metadata together with the input audio (GR-AudioLLM). Compared with the Audio-LLM, GR-AudioLLM consistently improves accuracy across all evaluated datasets, with particularly large gains on Vietnamese Common Voice (+60.00,pp), English FLEURS (+47.70,pp), Tamil Common Voice (+37.72,pp), and Malay SMALDUSC (+41.20,pp). These results demonstrate that explicitly providing gender information substantially enhances the Audio-LLM's ability to answer gender-related paralinguistic questions. This observation is consistent with our previous findings~\citep{cpqa_interspeech}, where injecting paralinguistic metadata similarly improved contextual paralinguistic question answering (CPQA), suggesting that explicit paralinguistic cues provide an effective conditioning signal for speech-language understanding.
\begin{table}[h]
\centering
\caption{Accuracy (\%) on public datasets.}
\label{tab:public}
\setlength{\tabcolsep}{3pt}
\begin{tabular}{llccc}
\toprule
\textbf{Lang.} & \textbf{Test Set} & \textbf{Vox-Profile} & \textbf{MERaLiON-GR} & \textbf{Audio-LLM} \\
\midrule
\multirow{3}{*}{English} & FLEURS & 99.69 & \textbf{100.00} & 49.61 \\
 & IEMOCAP & 97.31 & \textbf{98.90} & 97.21 \\
 & Common Voice & 92.60 & \textbf{93.90} & 52.10 \\
\midrule
Chinese & Common Voice & 96.10 & \textbf{98.10} & 64.80 \\
\midrule
Malay & SMALDUSC & \textbf{97.60} & 93.20 & 57.10 \\
\midrule
\multirow{3}{*}{Tamil} & OpenSLR & 98.30 & \textbf{100.00} & 53.70 \\
 & EmoTa & 94.44 & \textbf{99.15} & 69.98 \\
 & Common Voice & 92.30 & \textbf{94.00} & 51.60 \\
\midrule
\multirow{3}{*}{Thai} & THAI-SER & \textbf{89.32} & 87.23 & 79.05 \\
 & Thai Elderly & 96.57 & \textbf{100.00} & 72.18 \\
 & Common Voice & 96.52 & \textbf{97.86} & 59.97 \\
\midrule
Vietnamese & Common Voice & 96.08 & \textbf{99.22} & 36.08 \\
\midrule
\multirow{2}{*}{Indonesian} & IndoWave & 95.33 & \textbf{98.33} & 71.00 \\
 & Common Voice & 95.00 & \textbf{97.40} & 50.02 \\
\midrule
Khmer & FLEURS & 99.74 & \textbf{99.74} & 69.80 \\
\bottomrule
\end{tabular}
\end{table}

%%%%%%%%%%%%%%%%%%%%%%%%%%

\begin{table}[h]
\centering
\caption{Accuracy (\%) on public datasets with paralinguistic metadata.}
\label{tab:meta_audiollm}
\setlength{\tabcolsep}{3pt}
\begin{tabular}{llccc}
\toprule
\textbf{Lang.} & \textbf{Test Set} & \textbf{Audio-LLM} & \textbf{MERaLiON-GR} & \textbf{GR-AudioLLM} \\
\midrule
\multirow{2}{*}{English}
  & FLEURS         & 49.61 & \textbf{100.00} & 97.31\\
  & IEMOCAP            & 97.21 & \textbf{98.90}  & 97.97\\
\midrule
Malay
  & SMALDUSC           & 57.10 & 93.20 & \textbf{98.30}\\
\midrule
\multirow{2}{*}{Tamil}
  & EmoTa          & 69.98 & \textbf{99.15}  & 92.30\\
  & Common Voice             & 51.60 & \textbf{94.00}  & 89.32\\
\midrule
Thai  & Common Voice             & 59.97 & \textbf{97.86}  & 72.90\\
\midrule
Vietnamese
  & Common Voice             & 36.08 & \textbf{99.22}  & 96.08\\
\midrule
Khmer
  & FLEURS         & 69.80  & \textbf{99.74} & \textbf{99.74} \\
\bottomrule
\end{tabular}
\end{table}

\subsection{In-the-Wild Results}

To evaluate robustness in unconstrained environments, we conduct 2-second segment-level evaluations on our internal in-the-wild recordings, consisting of 10--30-second utterances in four Singapore languages (Singlish, Chinese, Malay, and Tamil). Table~\ref{tab:wild} shows that the proposed model consistently outperforms Vox-Profile across all evaluation sets.

The improvements are particularly evident on these short, in-the-wild utterances, with gains of up to 4.32,pp. The largest improvement is achieved on Malay (+4.32,pp), where Vox-Profile also exhibits its lowest baseline accuracy, followed closely by Tamil (+4.16,pp). These results demonstrate that our model provides robust gender predictions for fine-grained, segment-level pseudo-labeling and paralinguistic metadata generation, making it a practical replacement for external GR models.

\begin{table}[h]
\centering
\caption{Accuracy (\%) on in-the-wild datasets (10--30\,s segments).}
\label{tab:wild}
\setlength{\tabcolsep}{3.5pt}
\begin{tabular}{llcc}
\toprule
\textbf{Lang.} & \textbf{Test Set} & \textbf{Vox-Profile} & \textbf{ MERaLiON-GR} \\
\midrule
Chinese &  SG-ECMT-Chinese         & 89.25 & \textbf{91.85} \\
\midrule
Singlish & SG-ECMT-Singlish         & 88.45 & \textbf{92.10} \\
\midrule
Malay & SG-ECMT-Malay    & 86.56 & \textbf{90.88} \\
\midrule
Tamil & SG-ECMT-Tamil    & 90.57 & \textbf{94.73} \\
\bottomrule
\end{tabular}
\end{table}

\section{Summary}
We presented MERaLiON-GR, a speech gender recognition model for English and Southeast Asian languages. Extensive experiments on both manually curated Singapore datasets and public multilingual benchmarks demonstrate that MERaLiON-GR consistently outperforms existing baselines across diverse languages, recording conditions, and speech durations. Furthermore, incorporating the predicted gender as paralinguistic metadata significantly improves the performance of a general-purpose Audio-LLM on gender-related speech understanding tasks. These results demonstrate the continued importance of dedicated speech encoders with task-specific paralinguistic modeling for robust gender recognition and cross-lingual generalization.

\begin{ack}

This research/project is supported by the National Research Foundation, Singapore under its National Large Language Models Funding Initiative. Any opinions, findings, conclusions, or recommendations expressed in this material are those of the author(s) and do not reflect the views of the National Research Foundation, Singapore. The computational work for this article was fully performed on resources of the
National Supercomputing Centre (NSCC), Singapore (https://www.nscc.sg). 

The authors would also like to thank Nattadaporn Lertcheva, Nabilah Binte Md Johan,  Amudha Narayanan, Anitha Veeramani, Siti Maryam Binte Ahmad Subaidi, Siti Umairah Md Salleh, James Tan, and Komathee Veerappan to perform human annotations for the gender annotation task. 

\end{ack}
\bibliography{custom}

\section{MERaLiON Team (alphabetical order)}
\label{sec:team}
Aw Ai Ti, Chen Fang Yih Nancy, Chiu Ying Lay, 
Ding Yang,
He Yingxu,
Jiang Ridong,
%Li Jingtao,
%Liao Jingyi,
Liu Zhuohan,
Lu Yanfeng,
Ma Yi,
%Manas Gupta,
Muhammad Huzaifah Bin Md Shahrin,
Nabilah Binte Md Johan,
Nattadaporn Lertcheva,
%Pan Chunlei,
Pham Minh Duc,
Sailor Hardik Bhupendra,
%Siti Maryam Binte Ahmad Subaidi,
Siti Umairah Binte Mohammad Salleh,
Sun Shuo,
Tarun Kumar Vangani,
Wang Qiongqiong,
%Won Cheng Yi Lewis,
Wong Heng Meng Jeremy,
Wu Jinyang,
%Zhang Huayun,
Zhang Longyin
%Zou Xunlong
% \section*{References}
% References follow the acknowledgments in the camera-ready paper. Use unnumbered first-level heading for
% the references. Any choice of citation style is acceptable as long as you are
% consistent. It is permissible to reduce the font size to \verb+small+ (9 point)
% when listing the references.
% Note that the Reference section does not count towards the page limit.
% \medskip

% {
% \small

% [1] Alexander, J.A.\ \& Mozer, M.C.\ (1995) Template-based algorithms for
% connectionist rule extraction. In G.\ Tesauro, D.S.\ Touretzky and T.K.\ Leen
% (eds.), {\it Advances in Neural Information Processing Systems 7},
% pp.\ 609--616. Cambridge, MA: MIT Press.

% [2] Bower, J.M.\ \& Beeman, D.\ (1995) {\it The Book of GENESIS: Exploring
%   Realistic Neural Models with the GEneral NEural SImulation System.}  New York:
% TELOS/Springer--Verlag.

% [3] Hasselmo, M.E., Schnell, E.\ \& Barkai, E.\ (1995) Dynamics of learning and
% recall at excitatory recurrent synapses and cholinergic modulation in rat
% hippocampal region CA3. {\it Journal of Neuroscience} {\bf 15}(7):5249-5262.
% }

%%%%%%%%%%%%%%%%%%%%%%%%%%%%%%%%%%%%%%%%%%%%%%%%%%%%%%%%%%%%

% \appendix

% \section{Appendix / supplemental material}

% Optionally include supplemental material (complete proofs, additional experiments and plots) in appendix.
% All such materials \textbf{SHOULD be included in the main submission.}

%%%%%%%%%%%%%%%%%%%%%%%%%%%%%%%%%%%%%%%%%%%%%%%%%%%%%%%%%%%%

\end{document}